# Research Impact of Solar Panel Cleaning Robot on Photovoltaic Panel's Deflection

Trung Dat Phan[1,2], Minh Duc Nguyen[1,2], Maxence AUFFRAY[4], Nhut Thang Le[1,2], Cong Toai Truong[1,2], Van Tu Duong[1,2], Huy Hung Nguyen[3], Tan Tien Nguyen [1,2,*]

[1]*National Key Laboratory of Digital Control and System Engineering (DCSELab), Ho Chi Minh City University of Technology (HCMUT), 268 Ly Thuong Kiet Street, District 10, Ho Chi Minh City, Vietnam.*

[2]*Vietnam National University Ho Chi Minh City, Linh Trung Ward, Thu Duc City, Ho Chi Minh City, Vietnam.*

[3]*Faculty of Electronics and Telecommunication, Sai Gon University.*

[4]*Faculty of Mechanical Engineering, Higher Institute of Materials and Advanced Mechanics of Le Mans*

[*]*Corresponding author. E-mail: nttien@hcmut.edu.vn*

## Abstract

In the last few decades, solar panel cleaning robots (SPCR) have been widely used for sanitizing photovoltaic (PV) panels as an effective solution for ensuring PV efficiency. However, the dynamic load generated by the SPCR during operation might have a negative impact on PV panels. To reduce these effects, this paper presents the utilization of ANSYS software to simulate multiple scenarios involving the impact of SPCR on PV panels. The simulation scenarios provided in the paper are derived from the typical movements of SPCR observed during practical operations. The simulation results show the deformation process of PV panels, and a second-order polynomial is established to describe the deformed amplitude along the centerline of PV panels. This second-order polynomial contributes to the design process of a damper system for SPCR aiming to reduce the influence of SPCR on PV panels. Moreover, the experiments are conducted to examine the correlation between the results of the simulation and the experiment.

## 1 Introduction

In the 4.0 industrial revolution period, the human necessity to use energy is higher than ever before, and it is a big challenge for the energy industry in the world. In that context, many countries are conducting a transition from fossil fuels to renewable energy sources. Among the various renewable energy sources, solar energy has emerged as a solution for a sustainable future due to its cost-effectiveness, smaller installation space requirement, lower maintenance costs, and high performance compared to other energy sources[1]. In fact, the total installed solar panels have increased by nearly 650 GW [2] over the past decade, which is expected to increase by 1,100 GW in 2026 [3]. However, maintaining the performance of a photovoltaic (PV) panel system poses significant challenges due to various factors. One of the major issues is the accumulation of dust and debris on the surface of PV panels, which can lead to a decrease in efficiency of up to 30% [4]. Therefore, periodic cleaning of the PV panel system is necessary to ensure its efficiency.

Currently, there are numerous studies on cleaning PV panels that have shown many cleaning methods around the world, such as manual brushing, nanocoatings, chemicals, electrostatic force, ultrasonic waves, and automated approaches such as an automatic water spraying system or a robot with cleaning tools [5], [6]. Each method of cleaning PV panels has its advantages and disadvantages, which depend on factors such as the installation structure of the panels and the characteristics of dust in a particular area. Among these methods, the utilization of mobile robots is considered a popular solution, to date, due to its high level of automation, fast cleaning speed, and overall efficiency [7]. Nevertheless, the dynamic load generated by the cleaning robot when moving on the surface of the PV panel can experience deformation and induce microcracks in the silicon cells, which directly impact the panels' lifespan and performance [8]. Consequently, the panels degrade at a rate of approximately 0.8% to 1.1% per year [9], resulting in a decrease in photovoltaic efficiency of around 20% [10]. To solve the above problem, the application of damping mechanisms has become widely adopted in order to minimize the effects of robots on the PV panel during the cleaning process [11]. Besides the fact that there is limited research on the surface of PV panels, the design of the shock absorber assembly depends on the particular terrain where the robot operates. Thus, there are restricted studies on the design of damper systems for solar panel cleaning robots (SPCR).

This paper examines the deformation of PV panels caused by the weight of the SPCR and presents a polynomial describing the profile of the PV panel at maximum deformation. This polynomial describes the coordinates of the deformation points at the most dangerous cross-section, which are exerted by the wheel of the SPCR. The equation contributes to the design process for shock absorbers. The procedure to obtain the objective of this research is to follow as:

- The deformation of the PV panel is investigated by computational simulation through ANSYS based on the panel structure and parameters of SPCR.
- The polynomial describing the deformation of the PV panel at the cross-section exhibiting the highest amplitude is determined.
- The correlation between simulation studies and experimental results is achieved.

## 2 Overall System Description

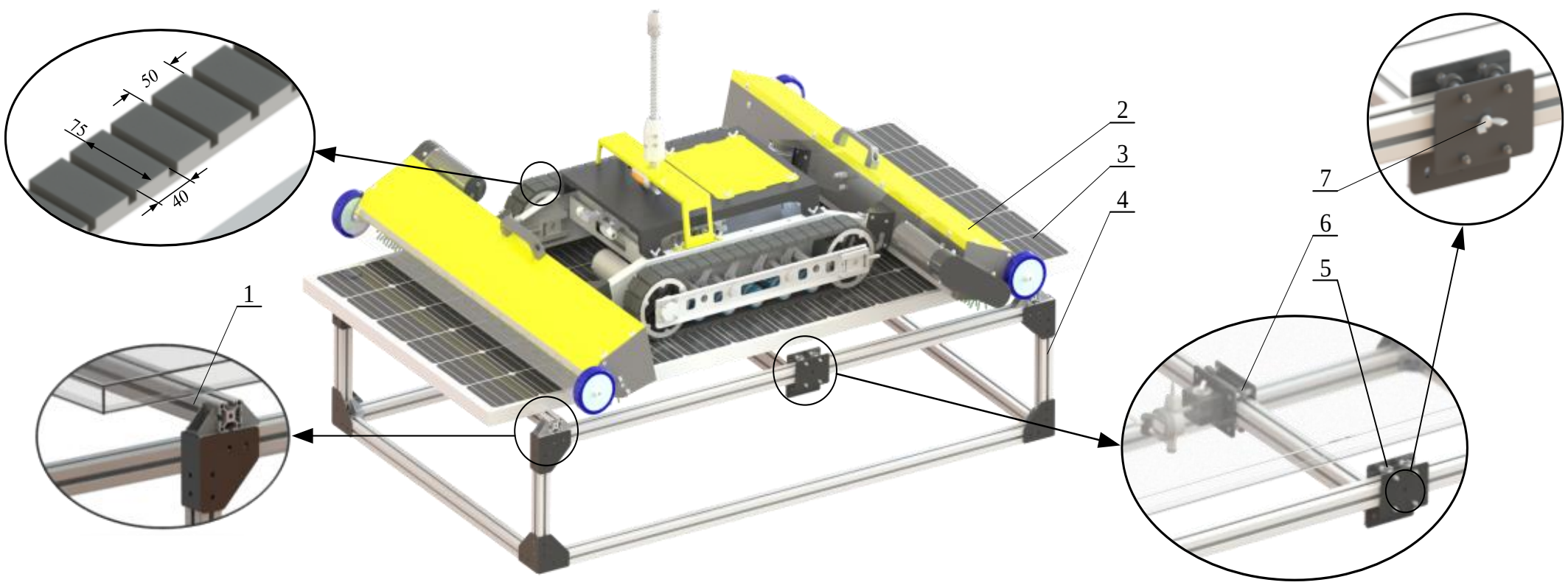

**Figure 1. 3D CAD of test bed.**

Figure 1 depicts the test bed to examine the deformation of the PV module under the load of the SPCR. The experimental system consists of five components: SPCR (2), PV panel (3), support frame (4), 2-axis sliding examine (5), and Mitutoyo 543-400B round-type dial gauge (6). SPCR functions as a load on the PV panel (3), with two-wheel assemblies being contact areas. The PV panel (3) is attached to the support frame (4) using two support bars (1), similar to the actual installation at PV power plants. Because the slippage is not considered in this research, the installation of PV panels is parallel to the ground without loss of generality. To measure the deformation, the round-type dial gauge is traversed across the PV panel thanks to a 2-axis sliding mechanism (5). To measure the deformation of PV panels, the locking nuts (7) are utilized to block the sliders at the location of measured deformation.

## 3 Material and Method

### 3.1 Physical model

In this simulation, the support bars are assumed to be perfectly rigid, with negligible deformation. Additionally, a PV panel has five adjacent layers in the sequence of protective cover (tempered glass), encapsulant (EVA), semiconductor (Silic), and back sheet (Tedlar or Polypropylene) [12]. Besides, these layers together form a compact and robust unit, firmly held within an aluminum frame. In these layers, the protective cover, made of tempered glass, serves as the main bearing component and protects the PV panel from external payloads. The EVA and Tedlar layers are thin film materials and do not greatly affect the bearing capabilities of the PV panel. The PV cell layer is the most sensitive and requires protection as it is prone to the formation of microcracks under vibrations, as discussed in Section 1. Therefore, to reduce computational time and simplify the physical model, only the tempered glass layer is investigated. For monocrystalline or polycrystalline PV panels, the standard thickness of the tempered glass layer ranges from 3-4mm [13], [14] and 3.5mm tempered glass was used for this simulation. Additionally, the aluminum frame is also another bearing element in a PV panel. According to Smartclima [15], there are various cross-section profiles available for aluminum frames, with up to 60 different profiles, but the TNY038 one was chosen for modeling the aluminum frame because it is compatible with the parameter of PV panel type RD320TU-36MD. Moreover, insignificant factors like fillets and tiny grooves in PV panels were ignored to avoid errors when meshing and calculating. Hence, based on the combination of some pre-processing conditions, the cross-sectional profile of the frame is shown in Figure 2.

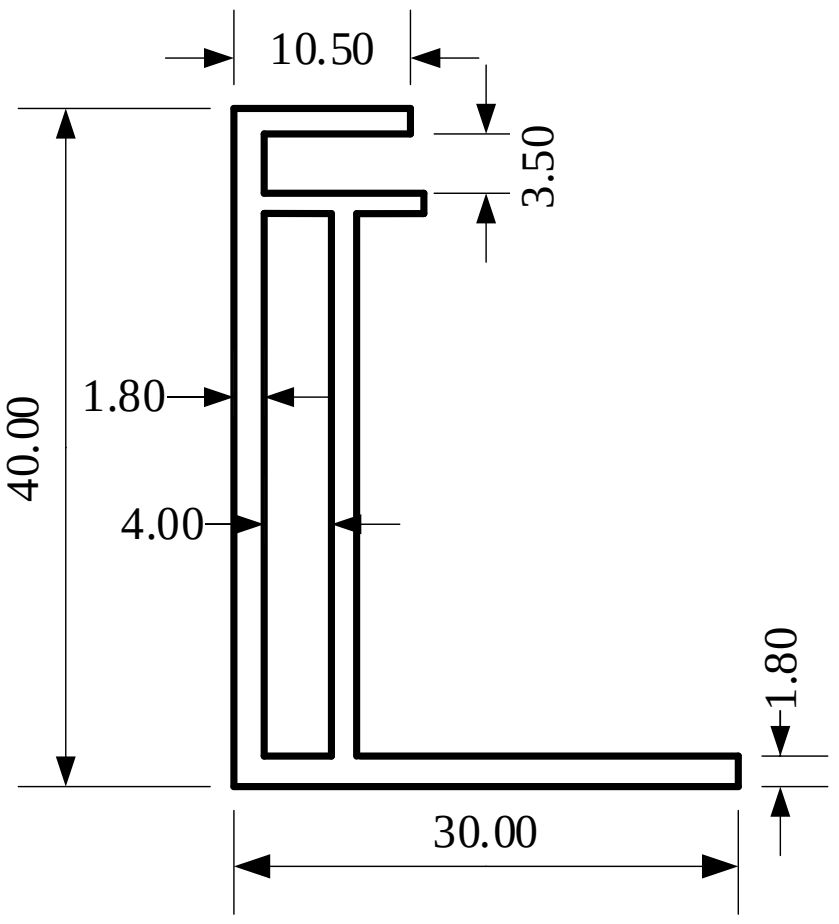

**Figure 2. The profile of aluminum frame cross-section.**

To simulate the deformation of a PV panel, three important parameters of the tempered glass layer need to be considered, including mass density, Young's modulus, and Poisson ratio [16]. Referring to previous studies [17], [18], the mechanical properties of aluminum frames and tempered glass are summarized in Table 1.

**Table1. Material properties for simulation**

| Materials | Young's modulus, GPa | Poisson ratio | Mass density, $kg.m^{-3}$ |
|---|---|---|---|
| 6063-T5 Aluminium | 70 | 0.3 | 2700 |
| Tempered Glass | 73 | 0.23 | 2500 |

According to [19], [20], the PV panels recommended for the installation are located approximately 200 to 400mm from the short edges of the aluminum frame. In addition, areas outside of this zone are considered hazardous where the PV panel might be subjected to wind loads, potentially causing damage. In some cases, the PV panel installation does not adhere to standards, so the PV panels can become highly susceptible to an external load. To assess the worst-case scenario, the clamps, which are the pads to fix between the PV panel and support frame, were positioned at a distance of 200mm from the short edge of the aluminum frame. Specifically, this is done to account for the potential risks and ensure a comprehensive evaluation of the system under adverse conditions. The investigation also revealed that the robot barely slips when operating on PV frames with an incline of 10° [21]. Therefore, in the static state, the SPCR with a mass of 83kg exerts a force of approximately 81.5kg on the PV panel surface. Besides, the boundary conditions for the simulation process are described in Table 2.

**Table2. Boundary conditions for the simulation process**

| Parameters | Value |
|---|---|
| Overall dimensions, mm | 1956×992×40 |
| Width of clamping area, mm | 40 |
| External load, kg | 81.5 |
| Length of contact area, mm | 590 |
| Distance between two belts, mm | 673 |

### 3.2 Simulation method

The finite element method (FEM) is used for strain analysis of PV panels, with the simulation process described in Figure 3. Initially, the mechanical properties shown in Table 1 were defined. Subsequently, a physical model of the PV panel was created using ANSYS SpaceClaim. The contact areas between the belt and the surface of the PV panel, as well as between the support bars and the aluminum truss, were also generated. The physical model of the PV panel was then discretized into smaller elements by ANSYS Meshing. The tempered glass sheet, with its simple structure, was meshed using 10 mm-hexahedral elements, while the aluminum frame was meshed using 5 mm-tetrahedral ones. This approach facilitates convergence and enhances result accuracy, considering the complexity of the aluminum frame compared to the tempered glass. Next, the boundary conditions for fixed support and external load were set according to the parameters mentioned in Table 2. Finally, the solution for directional deformation was defined, and the necessary data was extracted for determining the curvature deformation of the PV panel.

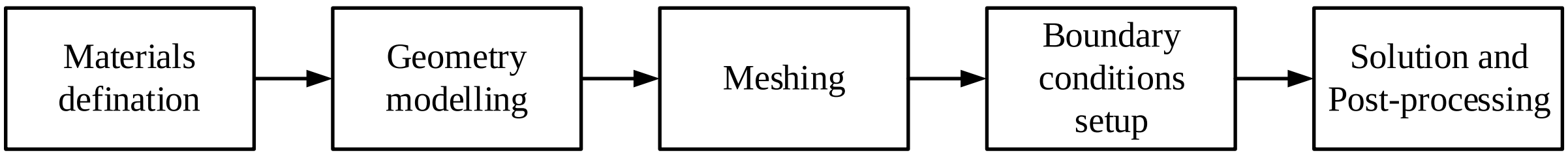

**Figure 3. Steps for a FEM simulation using Ansys software.**

The PV panel was investigated in two different trajectories of SPCR, as shown in Figure 4. These trajectories represent the typical SPCR motions across consecutive PV panels in practice. Hence, in the scope of this paper, the robot moves along the length of the PV panel chosen to be considered (Figure 4). The magnitude and position of the force applied to the PV panel vary depending on the position of the SPCR. Assuming the force created by SPCR's mass is evenly distributed on the belts, the force acting on the PV panel in any situation is proportional to the belt contact area.

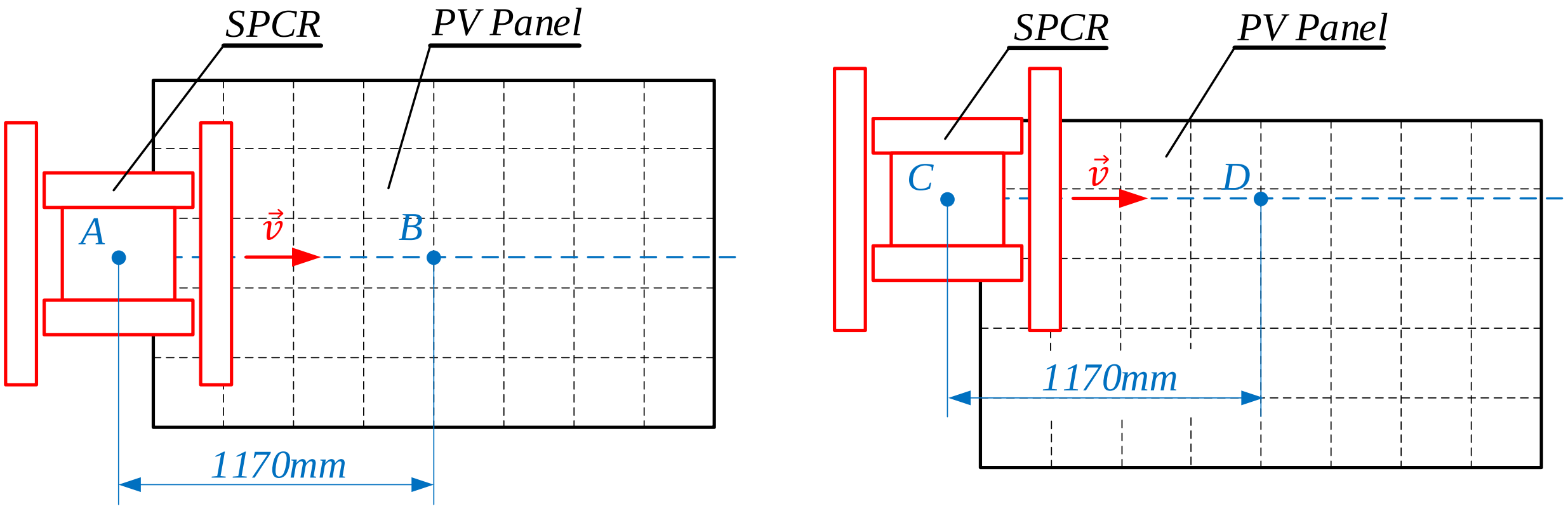

a), Central linear movement b), Side linear movement

**Figure 4. Simulation cases for robot movement on the surface of PV panels.**

Since the system is symmetric, only one-quarter of the panel dimensions were considered. The scenarios are described in detail in Table 3.

**Table3. Scenarios descriptions**

| Scenarios | Descriptions |
|---|---|
| Central linear movement | The SPCR was moved from point A to point B along the centerline of PV panels, as shown in Figure 4. The simulation was conducted at 10 positions, each spaced 120 mm apart. |
| Side linear movement | The SPCR's position was moved from point C to point D along to the sides of PV panels. |

## 4 Results and Discussion

### 4.1 Simulation result

Figure 5 shows the deformation contour of the PV panel in the two scenarios. In the first case as shown in Figure 5a), the SPCR moves closer to the center of the PV panel, and the deformation becomes increasingly pronounced due to the growing pressure exerted on the PV panel. The region of significant deformation is primarily focused beneath the center of gravity of the SPCR. The PV panel experiences a maximum displacement of 10.4 mm in this area, which afterward gradually diminishes towards the edges of the support frame. In the second as shown in Figure 5b), the contour shows the same trends as in the first case. However, there is a slight difference in the maximum displacement, that is, the region located near the contact area between the inner belt-based track of the SPCR and the PV panel surface shows the deformation of 11.3 mm.

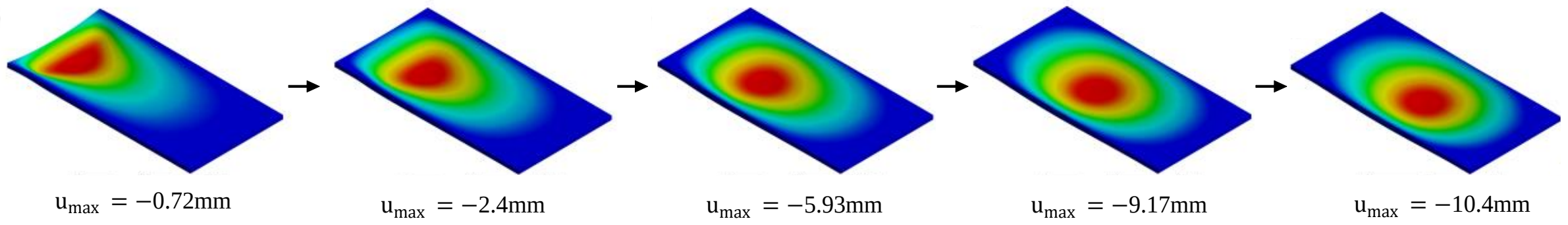

a), Displacement contours of the PV panel in the first scenario

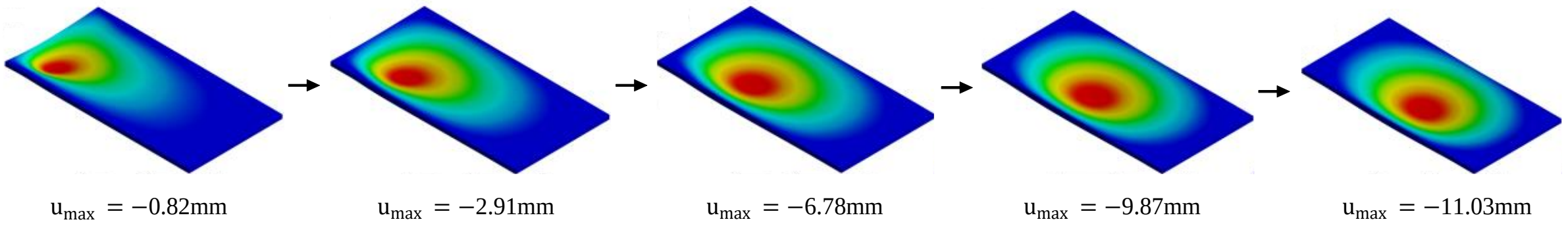

b), Displacement contours of the PV panel in the second scenario

**Figure 5. Simulation results of the central and slide linear movement cases.**

Based on the above results, the simulation of deformation at the center of the PV panel in the second case is further considered. Place the origin at the point lying on the short edge of the panel; the x-axis is parallel to the long side of the PV panel under non-deformed; the x-axis is perpendicular to the surface of the PV panel, in Figure 6. Equation (1), derived from the simulation results, characterizes the relationship between the deformation of the PV panel and its length.

$$y = 2.032\text{x}10^{-5}x^2 + 0.039x + 8.414 \quad (1)$$

where x represents the horizontal coordinate and y represents the deformation at that specific point.

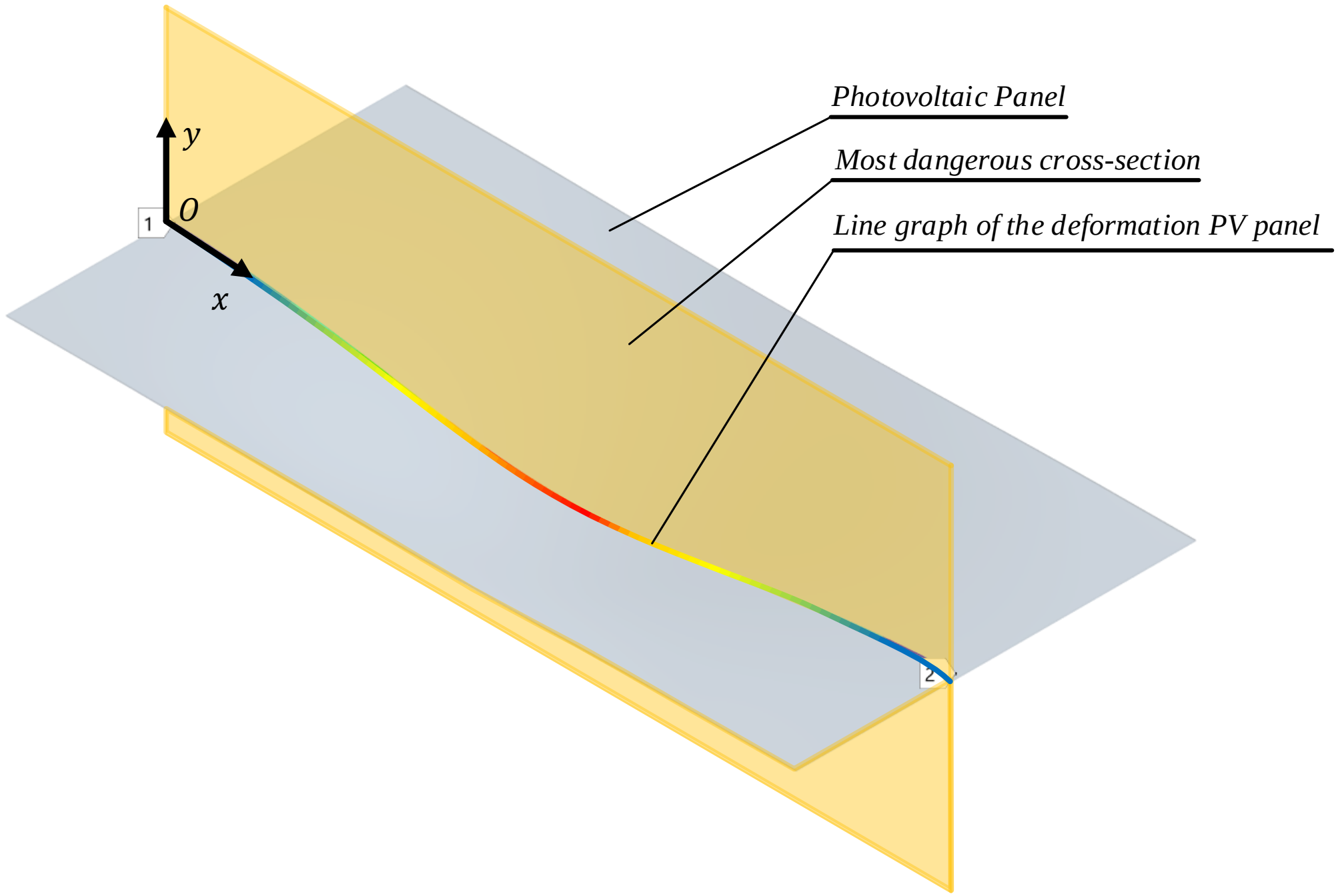

**Figure 6. Hazardous zone cross-sections of PV panels follow the result of the simulation.**

## 4.2 Experiment result

Figure 7 shows the experimental setup that examines the most dangerous position of the PV panel as concluded in subsection 4.1. In this study, the Mitutoyo 543-400B round-type dial gauge was employed, offering a precision of 0.01mm. Furthermore, considering the extensive data collection required for the PV panel in experiments, a systematic approach is adopted to maintain the consistency of the measurements. This approach consists of four sequential steps outlined in Table 4.

**Table4. Experimental steps to measure the deformation of the panel in the most dangerous case**

| Step | Descriptions |
|---|---|
| 1 | Zero-return the Mitutoyo 543-400B measuring device corresponding to the flat surface of the PV panel without the SPCR. |
| 2 | Place the SPCR at the position where the maximum deformation occurs according to the simulation. |
| 3 | Mount the measurement tool along the line graph of the deformation PV panel. |
| 4 | Proceed by systematically moving the measurement tool linearly along two designated points on the PV panel. Maintain a consistent increment of 10mm between each measurement, recording the deformation observed at each point. |

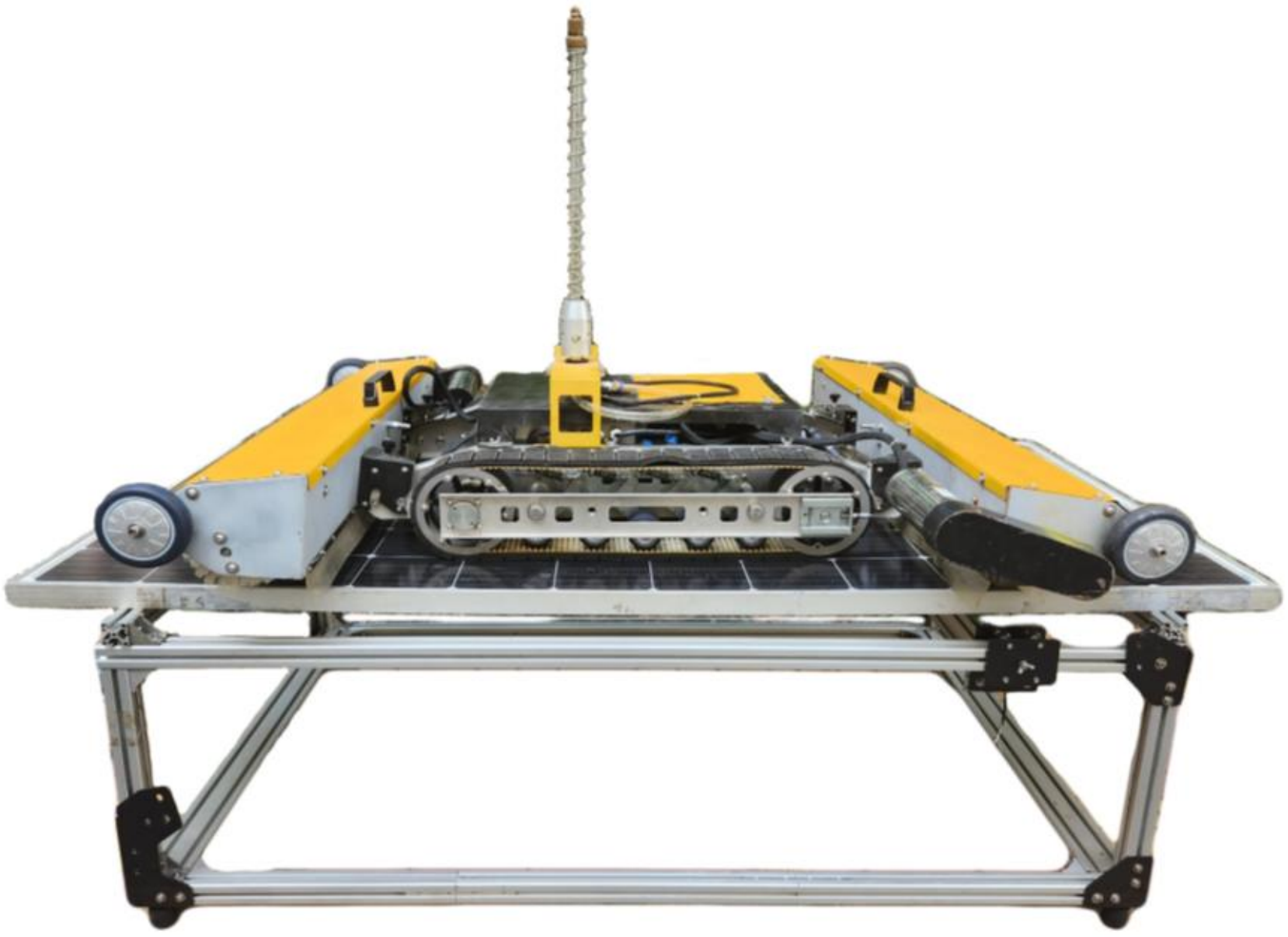

**Figure 7. Experimental system to measure the deformation of the PV panel in the most dangerous case.**

Figure 8 describes the line graph of the PV panel at the most dangerous cross-section in both simulation and experimental cases. The horizontal axis is the position where the round-type dial gauge is located, and the vertical axis is the deformation at that specific point. The experimental graph reveals that the deformation pattern of the PV panel, with the SPCR positioned at the middle of the right/left edge, follows a parabolic trend. The point of minimum deformation (-10.88mm), belonging to the parabolic trend, represents the location of the most pronounced deflection.

Figure 8 illustrates two-line graphs derived from simulation data and experimental data. To evaluate the accuracy of the simulation, the correlation coefficient between the deformation line graphs of the simulation data and the experimental data is calculated. According to [22], the correlation coefficient ($r$) between two-line graphs is determined as follows:

$$r = \frac{\sum_{i=1}^{n}(x_i - \bar{x})(y_i - \bar{y})}{(n-1)\sqrt{s_x^2 s_y^2}} \quad (2)$$

where $x_i$ and $y_i$ are the individual data points of the simulation and experimental data sets. $\bar{x}$ and $\bar{y}$ are the means of two datasets. $n = 126$ is the number of samples. $s_x^2$ and $s_y^2$ are the variances of two datasets, $a = \{x, y\}$, calculated as follows:

$$s_a^2 = \frac{\sum_{i=1}^{n}(a_i - \bar{a})^2}{n-1} \quad (3)$$

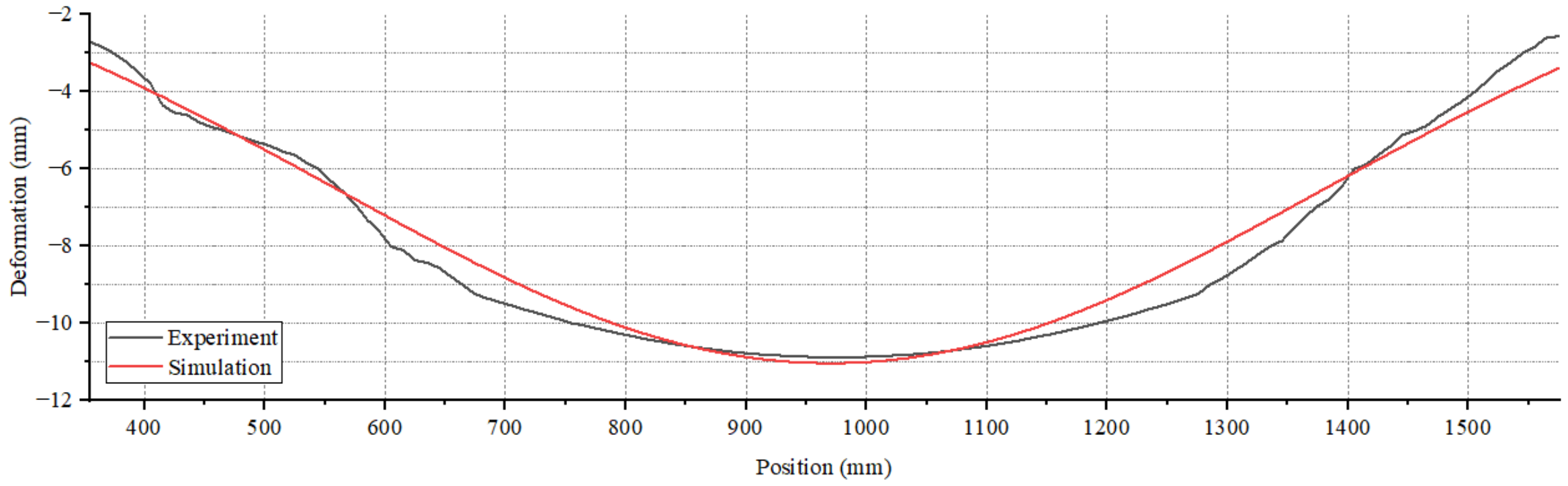

**Figure 8. Graph of PV panel at the most dangerous cross-section by experiment and simulation.**

Besides, the correlation coefficient calculated as r = 0.988 (close to +1), which shows experimental result data and simulation result data are positive strong linear relationships according to [22]. The experiment results show minimal deformation at position 975 mm, closely aligning with the simulation result (with an error of approximately 1.31%). In the line graph, the lowest point of deflection is important as it is the basis for designing the range of the working area of the damper system which aims to diminish the negative impact of the SPCR on PV panels. Notably, the experimental result (10.88mm) at the lowest point of deflection is lower than that of the simulation result (11.03 mm), so these conditions are completely valid for the calculations involved. Particularly considering that this study explores the most critical scenario for PV panel mountings (section 3), variations that result in lesser deformation of the panels do not diminish the significance of equation (1). It still remains relevant and valuable in informing the design of damper systems. With the given results, the initial simulation conditions and physical model in Section 3 are suitable. Moreover, the correlation serves as a confirmation of the underlying assumption that tempered glass and aluminum frames are primary load-bearing components of PV panels.

## 5 Conclusion

This paper studies the impact of the SPCR on PV panel deflection. A simulation model with appropriate boundary conditions was established to investigate the different deformations of the PV panel in two different scenarios. The results indicated that as the SPCR approached the center of the PV panel, the deformation of the PV panel increased gradually in both cases of central linear movement and side linear movement. Moreover, for maximum deformation of the PV panel, the research provided an equation that was assessed through experimental results with a reasonable difference. The experimental results also demonstrated the reliability of this estimated equation and provided a theoretical basis for the calculation of damper system designs in the future.

However, the paper has some limitations when ignoring the periodic vibrations caused by the rotating brush assembly of the SPCR. These vibrations can contribute to the degradation of the support frame, and the PV panel and induce microcracks. Hence, the impact of the brush assembly on the SPCR will be investigated through experimentation and simulation in the future. Additionally, the PV panel deformation in different mounting configurations will be studied to provide a more comprehensive understanding of the addressed issues.

## Acknowledgments

This research is funded by Vietnam National University Ho Chi Minh City (VNU-HCM) under grant number TX2023-20b-01. We acknowledge the support of time and facilities from National Key Laboratory of Digital Control and System Engineering (DCSELab), Ho Chi Minh City University of Technology (HCMUT), VNU-HCM for this study.